%% file: main.tex
\documentclass[conference]{IEEEtran}
\IEEEoverridecommandlockouts
\usepackage{subcaption} % for subfigures
\usepackage{orcidlink}
\usepackage[normalem]{ulem}
\usepackage{cite}
\input{style/packages}

\title{ RF Spectrogram Anomaly Detection with Quantum Kitchen Sinks: Architecture, Representation, and Hardware Validation}

\begin{document}

\author{
\IEEEauthorblockN{
Abdallah Aaraba\IEEEauthorrefmark{1}\IEEEauthorrefmark{3}\orcidlink{0000-0003-2980-6406},
Alexis Vieloszynski\IEEEauthorrefmark{2}\IEEEauthorrefmark{3}\orcidlink{0009-0002-9669-9620},
Remon Polus\IEEEauthorrefmark{1}\orcidlink{0000-0002-5527-0265},
Soumaya Cherkaoui\IEEEauthorrefmark{1}\orcidlink{0000-0001-6140-770X},
Ola Ahmad\IEEEauthorrefmark{2}\orcidlink{0000-0001-6854-9866}
}

\IEEEauthorblockA{\IEEEauthorrefmark{1}
Department of Computer and Software Engineering, Polytechnique Montréal, Montréal, QC, Canada \\
Email: abdallah.aaraba@ieee.org, remon.polus@polymtl.ca, soumaya.cherkaoui@polymtl.ca}

\IEEEauthorblockA{\IEEEauthorrefmark{2}
Thales cortAIx Labs, Montréal, QC, Canada \\
Email: \{alexis.vieloszynski, ola.ahmad\}@thalesgroup.com}

\IEEEauthorblockA{\IEEEauthorrefmark{3}
Contributed Equally}

%\IEEEauthorblockA{\IEEEauthorrefmark{3}
%Department of Computer and Software Engineering, Polytechnique Montréal, Montréal, QC, Canada \\
%Email: \{remon.polus, soumaya.cherkaoui\}@polymtl.ca}
}
% ^------------^------------^----Do not want these spaces!



\newcommand{\ola}[1]{\textcolor{red}{[\textbf{ola}: #1]}}
\newcommand{\soumaya}[2]{\textcolor{magenta}{[\textbf{soumaya}: #1]}}
\newcommand{\alexis}[1]{\textcolor{blue}{[\textbf{alexis}: #1]}}

\maketitle

\begin{abstract}
The broadcast nature of wireless channels exposes radio-frequency (RF) networks to anomalous and malicious transmissions, making anomaly detection a fundamental requirement for secure spectrum management.
Quantum Kitchen Sinks (QKS) offer a lightweight hybrid quantum feature map suitable for near-term quantum devices, yet their behavior on structured signal data remains poorly understood.
In this paper, we extend the standard QKS template with multi-depth data re-uploading and ring entanglement, and evaluate the resulting pipeline on controlled RF spectrogram anomaly detection.
We introduce a validation-locked five-stage ablation protocol that systematically separates the effects of shallow architecture, re-uploading depth, episode budget, input representation, and classical readout.
Across the completed benchmark, Discrete Cosine Transform (DCT) representations consistently dominate raw and Principal Component Analysis (PCA) inputs, moderate-depth entangled QKS configurations form the strongest operating regime, and QKS improves over matched classical direct-readout baselines across all evaluated representation-readout pairs on the held-out test set, with the best configuration reaching a test Area Under the Receiver Operating Characteristic curve (AUROC) of 0.8778 and a test F1 of 0.7995. The study bridges two levels of realism: real measured 
sub-6\,GHz cellular signals on the data side and real-device validation on the \textit{ibm\_quebec} Quantum Processing Unit (QPU) on the computing side, with AUROC deviations below 0.013 relative to simulation. These results provide a practical, reproducible framework for deploying QKS-based anomaly detection in wireless networks.\\

%we investigate the performance of QKS for controlled radio-frequency (RF) spectrogram anomaly detection using a fixed coreset and a two-stage ablation workflow.
%The study combines a QKS architecture sweep on a fixed Discrete Cosine Transform (DCT) representation with a representation sweep over raw, DCT, and Principal Component Analysis (PCA) inputs using the best Stage 1 configurations.
%\textcolor{blue}{The best completed configuration uses an entangled 8-qubit, 512-episode QKS on DCT128x128 features and reaches a test F1 score of 0.7967.
%Across the completed experiments, DCT representations consistently dominate raw and PCA inputs, entangled high-episode QKS configurations form the strongest operating regime, and smaller entangled models approach the best score with only a small F1 gap.
%The resulting contribution is a practical recipe, with clear limitations, for applying QKS tocontrolled RF spectrogram anomaly detection.

%These results offer practical insights for effective deployment of QKS-based anomaly detection in wireless networks.
% fixed \ola{Abstract to be improved and corrected.}
\end{abstract}


% \fi
\IEEEpeerreviewmaketitle

%------------ sections ------------
\input{sections/introduction}
\input{sections/Dataset}
\input{sections/methodology}
\input{sections/experiments}

\input{sections/conclusion}

% ------------ references ------------
% \newpage
\bibliographystyle{IEEEtran} 
\bibliography{references/references}

%------------ appendices ------------
% \newpage
% \appendices
% \section*{Appendices}
% \addcontentsline{toc}{section}{Appendices}% if you want "Appendices" in the TOC
% \input{appendices/proofs}

\end{document}

%% file: style/packages.tex
\usepackage[utf8]{inputenc} % allow utf-8 input
\usepackage[T1]{fontenc}    % use 8-bit T1 fonts
\usepackage{hyperref}
\hypersetup{colorlinks=true, linkcolor=blue}
\usepackage{url}            % simple URL typesetting
\usepackage{booktabs}       % professional-quality tabless
\usepackage{nicefrac}       % compact symbols for 1/2, etc.
\usepackage{microtype}      % microtypography
\usepackage{graphicx}       % graphics
\graphicspath{{media/}}     % organize your images and other figures under media/ folder

\usepackage{xcolor}
\usepackage{amsmath,amssymb,amsfonts,amsthm}
\usepackage{mathtools}
\usepackage{siunitx}
\usepackage{float}
\usepackage{tikz}
\usepackage{quantikz}
\usetikzlibrary{arrows.meta, backgrounds, calc, fit, positioning}

%% file: sections/introduction.tex
\section{Introduction}
\label{sec:introduction}
Spectrum sharing is a fundamental concept in cognitive radio systems, enabling multiple heterogeneous networks to coexist while efficiently utilizing shared radio-frequency (RF) spectrum resources that are inherently limited \cite{jiang2022intelligent}.
This paradigm relies on the open and dynamic nature of the radio spectrum, where wireless transmission propagates over inherently shared channels \cite{zou2016survey}.
However, this openness inherently exposes wireless environments to a wide range of disruptions, including both unintentional disturbances and deliberate attacks such as unauthorized transmissions and malicious interference \cite{ali2024rf}.
Common examples of these adversarial actions include jamming, sniffing, spoofing, and the injection of anomalous signals \cite{lichtman2016lte}.
Among these threats, jamming attacks are especially harmful, as adversaries intentionally transmit disruptive signals to interfere with legitimate communications and distort observed spectrum activity \cite{lichtman2016communications}.
Such interference reduces the signal-to-interference-plus-noise ratio (SINR), resulting in higher error rates, degraded communication quality, and possible system blockage \cite{pirayesh2022jamming}.
Furthermore, the rapid advancement of software-defined radio technologies has significantly lowered the barrier to accessing and deploying jamming devices, making these threats more widespread and accessible \cite{wang2020dynamic}.
Consequently, both civilian and military wireless systems require robust and reliable spectrum monitoring and anomaly detection mechanisms to ensure secure and resilient operation in the presence of such adversarial activities.

Anomaly detection techniques aim to identify both known and unknown interference by detecting deviations from normal signal behavior \cite{rajasegarar2008anomaly}.
Early conventional methods, such as energy detection, cyclostationarity, and matched filtering \cite{yucek2009survey}, rely heavily on expert knowledge.
Recent works leverage machine learning (ML) on spectrogram datasets for jamming detection \cite{krause2023digital}, using supervised approaches with labeled data \cite{wu2017jamming,xu2022neural} and unsupervised methods such as autoencoders and prediction-based models \cite{rajendran2019unsupervised,tandiya2018deep}, deep Convolutional Neural Networks (CNN)\cite{ren2019using}, as well as Generative Adversarial Network (GAN)-based frameworks \cite{zhou2021radio}.
However, these techniques often suffer from high computational complexity, slow convergence, and limited adaptability in dynamic environments.

%To address these challenges, \ola{Are you sure that QML currently addresses adaptability in dynamic environment better than DNNs? This should be supported by empirical evidence otherwise it is an overselling. I suggest removing this 1st sent.}
\begin{figure*}[t]
\centering
\includegraphics[width=0.9\textwidth]{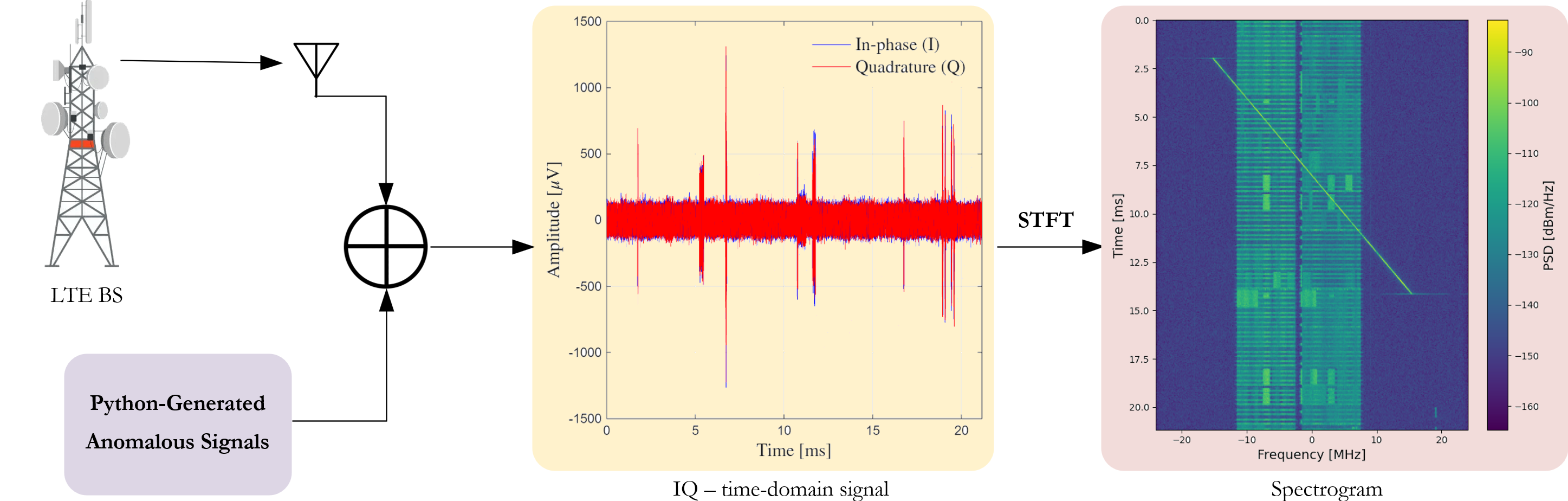}
\caption{Diagram of the data acquisition setup. It depicts the measurement of LTE signals, with simulated anomalous signals introduced into the environment. Subsequently, the measured IQ time-domain signals are transformed into spectrograms.}
%\ola{I suggest spreading all figures on one column to enhance the layout and reduce the extra empty space, last figure could be moved to the right of IQ figure.}}
\label{fig_01}
\end{figure*}

Quantum Machine Learning (QML) has emerged at the interface of quantum computing
and classical ML, with the goal of exploring whether quantum information
processing can enhance the efficiency or performance of learning
models~\cite{liao2025wireless, aaraba2024quack, vieloszynski2024latentqgan}. 
Building on this perspective, QML has also attracted growing interest for
anomaly detection in wireless systems, where high-dimensional signal
representations and complex interference patterns make efficient feature
extraction particularly important~\cite{ahmad2026quantum,aaraba2026randomized,hamhoum2025multivariate}. In particular,
Quantum Kitchen Sinks (QKS) adapt the classical random kitchen sinks framework
of explicit randomized feature maps to quantum circuits, thereby producing
finite-dimensional quantum features that can be processed by lightweight
classical readouts~\cite{wilson2018quantum}. QKS is especially attractive for
near-term noisy intermediate-scale quantum (NISQ) devices because the quantum
circuit acts as a randomized feature generator rather than a fully trainable
variational model, while the optimization burden is shifted to a simple
classical head~\cite{wilson2018quantum,cerezo2021variational}. This yields a
modular hybrid pipeline that is straightforward to train, compatible with
shallow circuit implementations, and particularly well suited to controlled
ablations across both input representation and circuit-design choices.
What remains missing in many application-oriented studies, however, is clear
characterization of which components of such a hybrid QKS pipeline matter most
when the inputs are structured signal representations such as RF spectrograms.
While QML holds significant potential to advance anomaly detection in RF
signals, its application to wireless spectrogram analysis remains largely
unexplored~\cite{mohammadisavadkoohi2025systematic}. To the best of our
knowledge, QKS has not yet been applied to anomaly detection in the wireless
spectrum, nor validated on real quantum hardware for this task. Motivated by
this gap, this paper presents an end-to-end QKS benchmark for controlled RF
spectrogram anomaly detection. We extend the standard shallow QKS
template~\cite{wilson2018quantum} with multi-depth data re-uploading and ring
entanglement, and evaluate the resulting pipeline through a validation-locked,
leakage-free, five-stage protocol that isolates shallow architecture,
re-uploading depth, episode budget, input representation, and classical readout
effects in sequence, without touching the test split until the final stage. Our
study bridges two levels of realism: real measured wireless signals on the data
side and real quantum hardware execution on the computing side, while anomaly
generation remains deliberately synthetic and controlled.

%\soumaya{ I modified the text above. I think the rest of text is too much detail for and introduction. The intro should motivate and state contributions, not describe methodology. Keeping it short here makes the contributions list hit harder immediately after. The above text should be followed by contributions}
%The study is built from a completed five-stage sklearn workflow \ola{Why sklearn is important here?}.
%It first searches shallow QKS architectures on a fixed mid-scale DCT representation, then tests whether repeated data re-uploading depth improves the shortlist \ola{Unclear}, checks budget-matched depth-versus-episode tradeoffs, transfers the strongest configurations across raw, DCT, and PCA representations, and finally evaluates matched flat-versus-QKS pairs on an untouched test split under multiple fast readouts.\\

The contributions of this paper are as follows.
\begin{itemize}
\item We construct a labeled spectrogram dataset combining real sub-6\,GHz cellular signals with synthetically generated anomalous signals spanning three interference types.
\item We extend the standard QKS template with multi-depth data re-uploading and ring entanglement, preserving a lightweight classical readout stage.
\item We introduce a validation-locked, leakage-free five-stage ablation protocol that separates shallow architecture, depth, episode budget, input representation, and readout effects without touching the test split until the final stage.
\item We show that the Discrete Cosine Transform (DCT) representations unlock strong QKS performance, with multi-depth QKS consistently improving over matched baselines across all evaluated readouts on the held-out test set.
\item Our study bridges two levels of realism by relying on real measured data and by validating the pipeline on the \textit{ibm\_quebec} Quantum Processing Unit (QPU), showing deviations in the Area Under the Receiver Operating Characteristic curve (AUROC) below 0.013 relative to simulation.
%\item we show that the strongest gains arise in DCT-based regimes, where multi-depth QKS features improve matched flat baselines on untouched test data while remaining readout-dependent and task-specific.
%\item \textcolor{blue}{Our study concludes with successful testing of the QKS algorithm on the IBM Quantum System One.}
\end{itemize}

%\ola{Check if QKS has not been implemented on real quantum machine, then this adds another contribution to the paper.} 
%\soumaya{I added it}

The remainder of this paper is organized as follows.
Section \ref{sec:Dataset} describes the generation of the spectrogram dataset used for both simulations and experiments.
Section \ref{sec:pipeline} introduces the proposed QKS architecture for anomaly classification.
Section \ref{sec:results} presents the performance evaluation of the proposed model.
Section \ref{sec:discussion} discusses the experimental results.
Finally, Section \ref{sec:conclusion} concludes the paper.

%% file: sections/Dataset.tex
\section{Spectrogram Dataset}
\label{sec:Dataset}

\begin{figure*}[!t]
    \centering
    
    % --------- Row 1: Main subfigures ---------
    \begin{subfigure}[t]{0.23\textwidth}
        \centering
        \includegraphics[width=\linewidth]{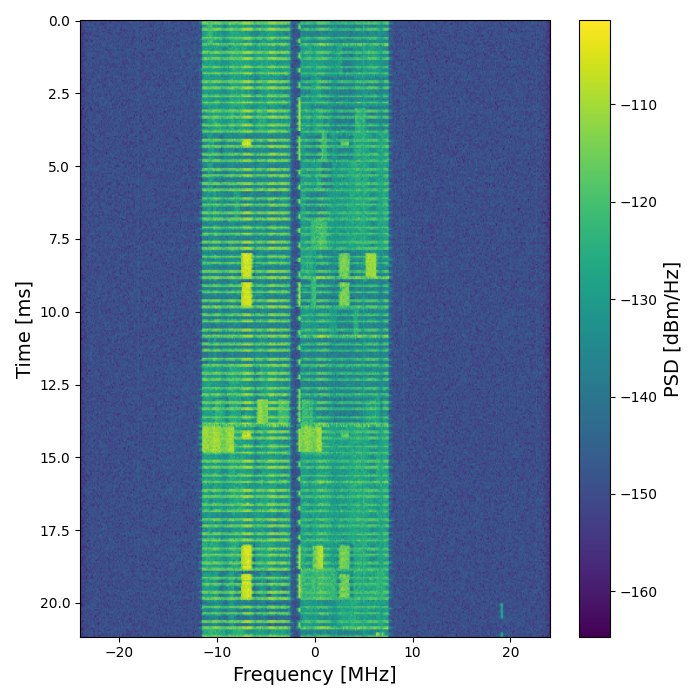}
        \caption{Normal Spectrogram}
        \label{fig:sub1}
    \end{subfigure}
    \hfill
    \begin{subfigure}[t]{0.23\textwidth}
        \centering
        \includegraphics[width=\linewidth]{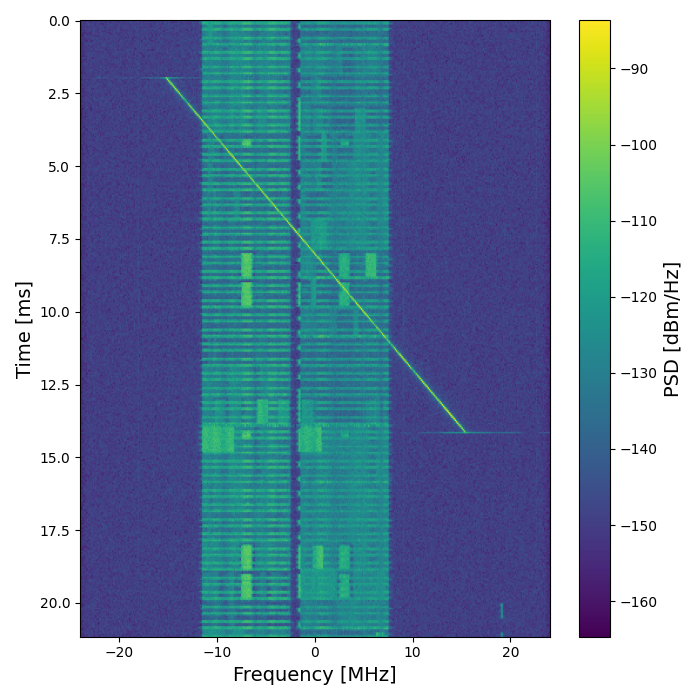}
        %\caption{Abnormal Spectrogram - Chirp}
        \caption{Chirp Anomaly}
        \label{fig:sub2}
    \end{subfigure}
    \hfill
    \begin{subfigure}[t]{0.23\textwidth}
        \centering
        \includegraphics[width=\linewidth]{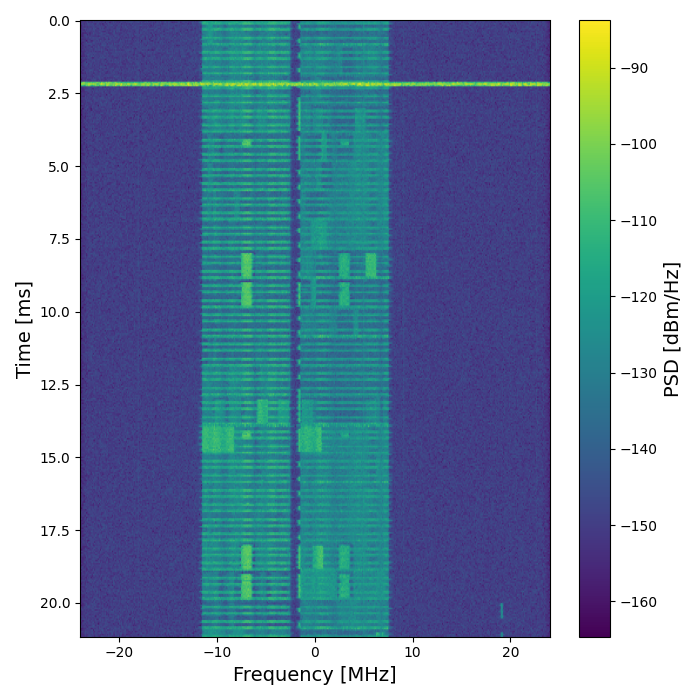}
        %\caption{Abnormal Spectrogram - Barrage Jamming}
        \caption{Barrage Jamming}
        \label{fig:sub3}
    \end{subfigure}
    \hfill
    \begin{subfigure}[t]{0.23\textwidth}
        \centering
        \includegraphics[width=\linewidth]{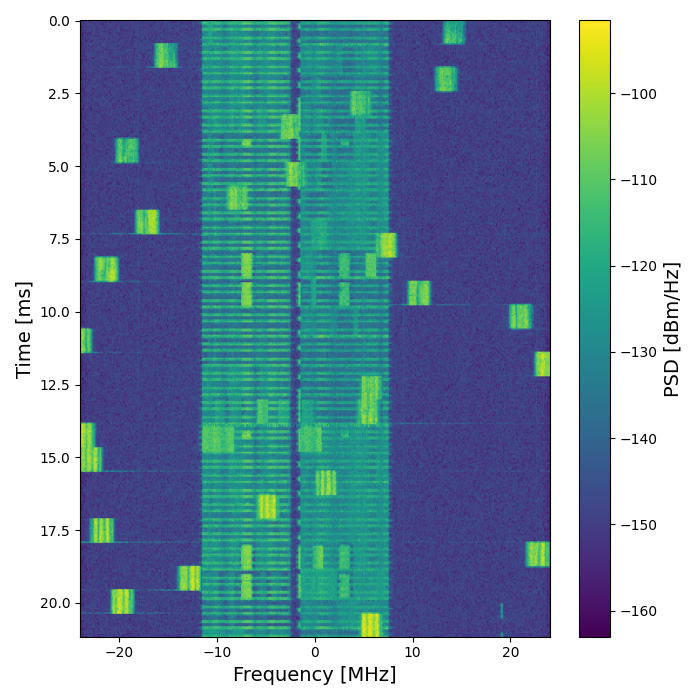}
        %\caption{Abnormal Spectrogram - Frequency Hopping Jamming}
        \caption{Frequency Hopping Jamming}
        \label{fig:sub4}
    \end{subfigure}
\caption{Example RF spectrograms from the dataset used in this study: (a) normal spectrogram; (b)–(d) spectrograms illustrating three different types of anomalies applied to the signal in (a).}
\label{fig_02}
\end{figure*}

This study leverages the dataset introduced in \cite{kim2024wireless}, which contains raw in-phase and quadrature (IQ) samples measured directly from real Long-Term Evolution (LTE) transmissions across multiple LTE frequency bands.
These real-world measurements serve as the primary user signals in cognitive radio networks, providing an authentic foundation for evaluating anomaly detection.
To simulate adversarial conditions, anomalous signals are synthetically generated using Python-based simulations and superimposed onto the measured LTE data (see Fig.\ref{fig_01}).
The analyzed sub-6 GHz bands are actively shared between LTE and 5G New Radio (NR) through dynamic spectrum sharing, making the detection of anomalous transmissions directly relevant to today's heterogeneous cellular networks.
%removed as per Ola's response to my question: %Furthermore, these same frequency ranges are utilized in tactical communications, where robustness against jamming and unauthorized signals is a critical operational requirement.

In this configuration, each data sample along the time axis corresponds to a duration of $21.15$ ms, while the frequency axis spans a bandwidth of $48$ MHz, as detailed in Table~\ref{Dataset_Summary}.
To emulate anomalous behavior within LTE bands, three types of interference signals are synthetically generated and embedded into the dataset.
Specifically, the chirp signal represents an unauthorized transmission \cite{kim2025spectrum}, while the other two—barrage jamming and frequency-hopping noise jamming (a time-varying form of partial-band interference)—correspond to intentional jamming activities \cite{lohan20245g}.
The temporal duration of these signals is randomly selected within $21.15$ ms, while their frequency locations are randomly distributed within the $48$ MHz band to ensure diversity.
The workflow of the dataset is illustrated in Fig.~\ref{fig_01}.
The embedding process is controlled by the Jamming-to-Signal Ratio (JSR), which quantifies the relative power of the injected anomalies with respect to the LTE signal.
Specifically, JSR values ranging from $-10$ dB to $5$ dB, with increments of $2$ dB, are considered to model varying interference conditions. The training set consists of $10{,}800$ unique LTE signals, and the test set consists of a disjoint set of $4{,}062$ signals. Since each signal is included in both its normal form and its anomaly-injected form, this yields the $21{,}600$ training samples and $8{,}124$ test samples reported in Table~\ref{Dataset_Summary}.

By transforming raw IQ samples into spectrograms, both legitimate transmissions and anomalous activities can be distinguished through their temporal and spectral characteristics.
Specifically, a time-domain signal of length $1{,}300{,}000$ IQ points is formed by combining the original LTE signal with the injected anomalies.
Subsequently, we generate the spectrograms by applying a Short-Time Fourier Transform (STFT) using a Hann window, with STFT parameters of $3250$ and $8192$, and an overlap ratio of $25\%$ \cite{kim2015low}.
The resulting time–frequency representation is subsequently downsampled to a resolution of $400 \times 400$.
Instead of applying a conventional Fast Fourier Transform (FFT) over the entire one-dimensional IQ sequence, the STFT partitions the signal into overlapping segments and computes the Fourier transform on each windowed portion.
These localized spectra are then concatenated along the temporal dimension to form a comprehensive time–frequency representation.
This approach facilitates the extraction of distinctive signatures associated with chirp, barrage, and frequency hopping patterns, enabling their differentiation from standard LTE activity.
The generated spectrograms are subsequently utilized as inputs to the QKS pipeline for anomaly detection.

\begin{table}[ht]
\centering
\fontsize{10}{12}\selectfont % Adjust font size to 8pt with line skip 10pt
\caption{Dataset Summary}
\label{Dataset_Summary}
\begin{tabular}{|c|c|}
\hline
Train Data samples & $21{,}600$ \\
\hline
Test Data samples & $8{,}124$ \\
\hline
Sampling Frequency & $61.44$ MHz   \\
\hline
Number of IQ points  & $1,300,000$  \\
per Data sample & \\ \hline
Time Duration & $21.15$ ms  \\
\hline
Bandwidth & $48$ MHz   \\
\hline
Spectrogram Size & $400\times400$   \\
\hline
Anomaly Signal & Chirp  \\
Classes & Barrage Jamming  \\
 & Frequency Hopping Jamming  \\
\hline
JSR Values [dB] & $[-10,-8,-6,-4,-2,0,2,5]$   \\
\hline
\end{tabular}
\end{table}

%\ola{It would be good to mention that from the normal IQ data provided, we generate our custom jamming signals.}
%\ola{Instead of saying spectrogram is generated, emphasis with we generated so it does not interpret as we downloaded the data as is, this also adds more sense in providing details on the dataset.}
An example of the generated dataset is illustrated in Fig. \ref{fig_02}.
The spectrogram of the normal LTE signal, shown in Fig. \ref{fig:sub1}, exhibits a structured and relatively stable pattern, where the signal energy remains confined within specific frequency bands corresponding to legitimate transmissions.
In contrast, the anomalous cases introduce noticeable distortions in the time–frequency representation.
As depicted in Fig. \ref{fig:sub2}, the chirp signal manifests as a continuous diagonal trace, representing an unauthorized transmission with its time-varying frequency.
In Fig. \ref{fig:sub3}, the barrage jamming spectrogram displays a wideband and dense energy distribution spanning the 48 MHz, indicating simultaneous interference across the entire bandwidth.
Meanwhile, Fig. \ref{fig:sub4} illustrates the frequency-hopping noise jamming, which is characterized by intermittent bursts that rapidly shift across several subbands, resulting in a scattered and discontinuous pattern.

%% file: sections/methodology.tex
\section{QKS Pipeline and Ablation Protocol}
\label{sec:pipeline}

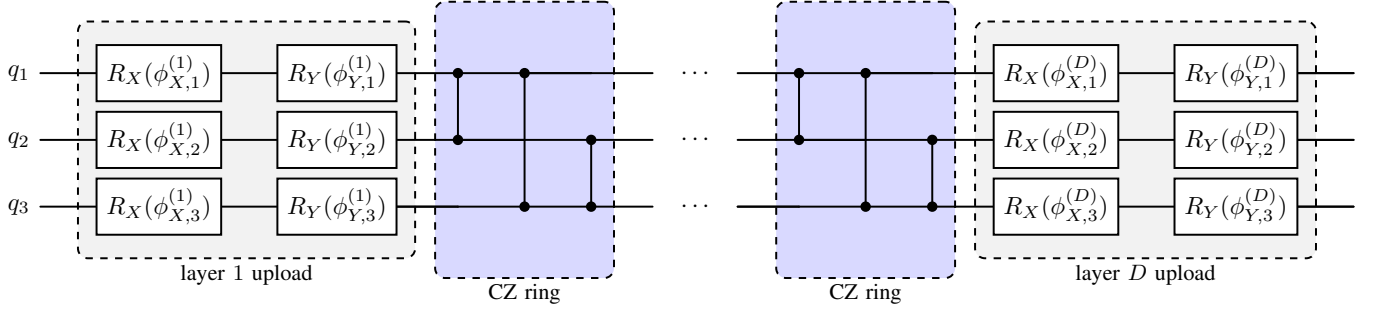
\begin{figure*}[t]
    \centering
    \resizebox{\linewidth}{!}{%
    \begin{quantikz}[row sep={0.95cm,between origins}, column sep=0.8cm]
        \lstick{$q_1$}
            & \gate{R_{X}(\phi_{X,1}^{(1)})}
            \gategroup[wires=3, steps=2,
                style={dashed, rounded corners, fill=gray!10, inner xsep=4pt, inner ysep=6pt},
                background,
                label style={label position=below, anchor=north, yshift=-0.15cm}
            ]{\small layer $1$ upload}
            & \gate{R_{Y}(\phi_{Y,1}^{(1)})}
            & \ctrl{1}
            \gategroup[wires=3, steps=3,
                style={dashed, rounded corners, fill=blue!15, inner xsep=4pt, inner ysep=14pt},
                background,
                label style={label position=below, anchor=north, yshift=-0.15cm}
            ]{\small CZ ring}
            & \ctrl{2}
            & \qw
            & \push{\hspace{4mm}\cdots\hspace{4mm}}
            & \ctrl{1}
            \gategroup[wires=3, steps=3,
                style={dashed, rounded corners, fill=blue!15, inner xsep=4pt, inner ysep=14pt},
                background,
                label style={label position=below, anchor=north, yshift=-0.15cm}
            ]{\small CZ ring}
            & \ctrl{2}
            & \qw
            & \gate{R_{X}(\phi_{X,1}^{(D)})}
            \gategroup[wires=3, steps=2,
                style={dashed, rounded corners, fill=gray!10, inner xsep=4pt, inner ysep=6pt},
                background,
                label style={label position=below, anchor=north, yshift=-0.15cm}
            ]{\small layer $D$ upload}
            & \gate{R_{Y}(\phi_{Y,1}^{(D)})}
            & \qw \\
        \lstick{$q_2$}
            & \gate{R_{X}(\phi_{X,2}^{(1)})}
            & \gate{R_{Y}(\phi_{Y,2}^{(1)})}
            & \control{}
            & \qw
            & \ctrl{1}
            & \push{\hspace{4mm}\cdots\hspace{4mm}}
            & \control{}
            & \qw
            & \ctrl{1}
            & \gate{R_{X}(\phi_{X,2}^{(D)})}
            & \gate{R_{Y}(\phi_{Y,2}^{(D)})}
            & \qw \\
        \lstick{$q_3$}
            & \gate{R_{X}(\phi_{X,3}^{(1)})}
            & \gate{R_{Y}(\phi_{Y,3}^{(1)})}
            & \qw
            & \control{}
            & \control{}
            & \push{\hspace{4mm}\cdots\hspace{4mm}}
            & \qw
            & \control{}
            & \control{}
            & \gate{R_{X}(\phi_{X,3}^{(D)})}
            & \gate{R_{Y}(\phi_{Y,3}^{(D)})}
            & \qw
    \end{quantikz}
    }
    \caption{Illustrative $3$-qubit multi-depth QKS episode schematic showing only the first and final layer-specific $R_{X}$-then-$R_{Y}$ uploads. CZ-ring entanglers are inserted only between successive upload layers, so the displayed pattern is $\text{layer }1\text{ upload} \rightarrow \text{CZ ring} \rightarrow \cdots \rightarrow \text{CZ ring} \rightarrow \text{layer }D\text{ upload}$. The special case $D=1$ recovers the shallow template.}
    \label{fig:qks_protocol_episode}
\end{figure*}

\subsection{Input representations}
Let $x \in \mathbb{R}^{400 \times 400}$ denote an RF spectrogram. The QKS pipeline does not operate directly on
this matrix, but rather on a vector representation $r(x) \in \mathbb{R}^{d}$ constructed
from it. In this study, we consider three such representation families:
\begin{align}
    r_{\mathrm{raw}}(x) &= \operatorname{vec}(x), \\
    r_{\mathrm{DCT}}(x) &= \operatorname{vec}\!\big(C_{1:k_f,\,1:k_t}(x)\big), \\
    r_{\mathrm{PCA}}(x) &= P_m\operatorname{vec}(x).
\end{align}
Here, $\operatorname{vec}(\cdot)$ flattens its matrix argument into a vector,
$C_{1:k_f,\,1:k_t}(x)$ denotes the upper-left $k_f \times k_t$ coefficient block of the 2-D DCT of $x$, and $P_m$ denotes the projection onto the first $m$ principal
directions learned from the training set.

The raw representation retains the full $400 \times 400$ spectrogram and
flattens it into a vector in $\mathbb{R}^{160000}$. The DCT family first applies a separable 2-D DCT (type II, with orthonormal normalization)~\cite{wallace_jpeg_1992} to the spectrogram, retains the upper-left $k_f \times k_t$ block of DCT coefficients, normalizes that block using
training-derived statistics, and then flattens it. This retained block corresponds to low-index DCT coefficients, that is, coefficients capturing coarse, slowly varying structure over the 2-D spectrogram array along both the frequency-bin axis and the time-bin axis, rather than low RF frequencies in the original signal.

In the benchmark, we use
$k_f \times k_t \in \{16 \times 16, 64 \times 64, 128 \times 128, 256 \times
256\}$. The Principal Component Analysis (PCA) family instead first flattens the spectrogram and then projects
it onto an $m$-dimensional linear subspace, with
$m \in \{32, 128, 256, 512, 2048\}$. These representation families compress the same signal in qualitatively
different ways. DCT is a fixed transform that explicitly preserves coarse 2-D spectrogram structure, while PCA is a learned linear projection applied to flattened spectrograms and preserves high-variance directions without maintaining the original grid structure in the same explicit way.
%DCT is a fixed transform that preserves coarse 2-D structure in the spectrogram by retaining only the lowest-index DCT coefficients along the frequency-bin and time-bin axes before flattening. PCA, by contrast, is a learned linear projection applied to already flattened spectrograms; it preserves directions of large empirical variance, but does not preserve the original 2-D grid structure in the same explicit way. 
This distinction is central to the
ablation: raw and DCT inputs become vectors only after flattening, whereas PCA
already returns a vector representation.

\subsection{Multi-depth QKS featurizer}
Given a representation vector $r(x) \in \mathbb{R}^{d}$, the QKS featurizer constructs one quantum feature block per episode. In line with the original QKS, multiple episodes are used to map the same input to different randomized quantum states, and each such per-episode state is measured through a fixed observable family to produce a corresponding per-episode feature block. The shallow QKS template~\cite{wilson2018quantum} corresponds to a single data-upload layer. Here, we extend that template to a depth-$D$ feature map with repeated data re-uploading.

For each episode $e \in \{1,\dots,E\}$ and layer
$\ell \in \{1,\dots,D\}$, the model samples two affine maps from the
representation space to the $n$ input-dependent circuit angles,
\begin{align}
    \phi_{X}^{(e,\ell)}(x) &= W_{X}^{(e,\ell)} r(x) + b_{X}^{(e,\ell)}, \\
    \phi_{Y}^{(e,\ell)}(x) &= W_{Y}^{(e,\ell)} r(x) + b_{Y}^{(e,\ell)},
\end{align}
with $W_{X}^{(e,\ell)}, W_{Y}^{(e,\ell)} \in \mathbb{R}^{n \times d}$ and
$b_{X}^{(e,\ell)}, b_{Y}^{(e,\ell)} \in \mathbb{R}^{n}$. Equivalently, for each qubit
$j \in \{0,\dots,n-1\}$,
\begin{align}
    \phi_{X,j}^{(e,\ell)}(x)
    &= \langle w_{X,j}^{(e,\ell)}, r(x) \rangle + b_{X,j}^{(e,\ell)}, \\
    \phi_{Y,j}^{(e,\ell)}(x)
    &= \langle w_{Y,j}^{(e,\ell)}, r(x) \rangle + b_{Y,j}^{(e,\ell)}.
\end{align}
In the completed benchmark, every entry of $W_{X}^{(e,\ell)}$ and
$W_{Y}^{(e,\ell)}$ is sampled i.i.d. from $\mathcal{N}(0,\sigma^2)$ with
$\sigma = 2$, which was found empirically to yield good results, while every entry of $b_{X}^{(e,\ell)}$ and $b_{Y}^{(e,\ell)}$ is sampled i.i.d. from $\mathcal{U}([0,2\pi])$.

These angle vectors parameterize a layered circuit template. We define the
single-layer data-upload operator as
\begin{equation}
    U_{e,\ell}(x)
    =
    \prod_{j=0}^{n-1}
    R_{Y}\!\left(\phi_{Y,j}^{(e,\ell)}(x)\right)
    R_{X}\!\left(\phi_{X,j}^{(e,\ell)}(x)\right),
\end{equation}
and we denote by $V_{\mathrm{ring}}$ either the identity (when entanglement is
disabled) or the fixed controlled-Z (CZ) ring
$(0,1),(1,2),\dots,(n-2,n-1),(n-1,0)$ otherwise. For episode $e$, the input state
$\lvert 0 \rangle^{\otimes n}$ is mapped to
\begin{equation}
    \lvert \psi_e(x) \rangle = U_e(x)\lvert 0 \rangle^{\otimes n},
\end{equation}
where
\begin{equation}
    U_e(x)
    =
    U_{e,D}(x)\,V_{\mathrm{ring}}\cdots
    U_{e,2}(x)\,V_{\mathrm{ring}}\,U_{e,1}(x).
\end{equation}
In other words, each episode re-uploads the same classical representation of the input across multiple randomized layers~\cite{perezsalinas2020datareuploading, schuld2021effect}, while entanglement, when enabled, is inserted only between consecutive upload layers. This preserves a lightweight per-layer circuit structure while allowing the effective quantum feature map to become richer as depth increases. In the circuit schematic below, each upload block is drawn left-to-right as an $R_{X}$ gate followed by an $R_{Y}$ gate, which corresponds to the operator product $R_{Y} R_{X}$ written above. The special case $D=1$ recovers the standard shallow QKS
template~\cite{wilson2018quantum}. Figure~\ref{fig:qks_protocol_episode} summarizes this multi-depth circuit structure.

In our method, the measured observable family is taken to consist of all one- and two-body Pauli observables,
\begin{equation}
\begin{aligned}
\mathcal{O}_n
= {} & \Bigl\{
    \sigma_a^{(i)}
    :
    0 \le i < n,\;
    a \in \{X,Y,Z\}
\Bigr\}
\\[2pt]
& \cup
\Bigl\{
    \sigma_a^{(i)}\sigma_b^{(j)}
    :
    0 \le i < j < n,\;
    a,b \in \{X,Y,Z\}
\Bigr\},
\end{aligned}
\end{equation}
so that $|\mathcal{O}_n| = 9\binom{n}{2}+3n$, where the factor $9$ comes from the $3 \times 3$ possible choices of $(a,b) \in \{X,Y,Z\}$ for each qubit pair $(i,j)$. This observable family captures both single-qubit statistics and pairwise correlations induced by the feature map, while remaining polynomial in $n$. The per-episode feature block is therefore
\begin{equation}
    \xi_e(x)
    =
    \big(\langle \psi_e(x) \rvert O \lvert \psi_e(x) \rangle\big)_{O \in \mathcal{O}_n}
    \in \mathbb{R}^{|\mathcal{O}_n|},
\end{equation}
and the final QKS feature vector is obtained by concatenating all per-episode feature blocks as follows,
\begin{equation}
    \xi(x)
    =
    [\xi_1(x)^\top,\dots,\xi_E(x)^\top]^\top
    \in \mathbb{R}^{E|\mathcal{O}_n|}.
\end{equation}
The number of episodes $E$ controls how many independently sampled feature
blocks contribute to $\xi(x)$, while the depth $D$ controls how many
data re-uploading and entanglement layers are applied within each episode.

\subsection{Five-stage protocol}
The benchmark is designed to disentangle the effects of architecture, depth,
depth--episode allocation, input representation, and final readout, while
keeping the raw test split untouched until the final evaluation. Stages~1--4
operate only on a fixed model-selection subset drawn from the raw training split
together with a disjoint validation subset. Stage~5 retrains the finalists on
the full raw training split and evaluates them once on the untouched raw test
split. The completed benchmark therefore contains
$60 + 24 + 4 + 20 + 30 = 138$ evaluations. For the benchmark reported here, the
model-selection subset used in Stages~1--4 comprises
approximately 20\% of the full training split, obtained by stratified random
sampling while preserving coverage across abnormal
groups; only Stage~5 uses the full training split. This reduced subset keeps
the search tractable without collapsing anomaly diversity.

\paragraph{Readouts}
During Stages~1--4, we deliberately use a single fast linear Support Vector Machine (SVM) readout so
that model selection remains focused on the quality of the quantum feature map
itself, rather than being confounded by a broader comparison of downstream
classifiers. In Stage~5, we then compare alternative corresponding readout families, as they
probe different aspects of the same QKS feature geometry, and evaluate their performance under a matched setting. A linear SVM tests
large-margin linear separability \cite{cortes_vapnik_1995}, logistic
regression tests whether the same feature geometry supports a stable
probabilistic linear decision rule \cite{cox_1958_binary}, and the Random Fourier Features (RFF) and
Nystr\"om lifts test whether additional gains appear only after a controlled
approximate-kernel nonlinearity is applied
\cite{rahimi_recht_2007_random_features,
williams_seeger_2001_nystrom}. This distinction tells us whether the QKS
features are already useful under simple linear models or whether their benefit
emerges only after a richer nonlinear readout is added. The shortlisted
representations are therefore evaluated with five matched readout families. For
any Stage~5 readout $\rho$, let
$f_\rho : \mathbb{R}^{E|\mathcal{O}_n|} \to \mathbb{R}$ denotes its score on
$\xi(x)$. We write
\begin{align}
    f_{\mathrm{linSVM}}(\xi)
    &= w_{\mathrm{svm}}^\top \xi + b_{\mathrm{svm}}, \\
    f_{\mathrm{log}}(\xi)
    &= w_{\mathrm{log}}^\top \xi + b_{\mathrm{log}}, \\
    f_{\mathrm{Rff\text{-}SVM}}(\xi)
    &= w_{\mathrm{rff\text{-}svm}}^\top \varphi_{\mathrm{Rff}}(\xi)
    + b_{\mathrm{rff\text{-}svm}}, \\
    f_{\mathrm{Rff\text{-}log}}(\xi)
    &= w_{\mathrm{rff\text{-}log}}^\top \varphi_{\mathrm{Rff}}(\xi)
    + b_{\mathrm{rff\text{-}log}}, \\
    f_{\mathrm{Nys\text{-}log}}(\xi)
    &= w_{\mathrm{nys}}^\top \varphi_{\mathrm{Nys}}(\xi) + b_{\mathrm{nys}}.
\end{align}
Here, $\varphi_{\mathrm{RFF}}$ and $\varphi_{\mathrm{Nys}}$ denote fixed
approximate-kernel feature lifts applied to $\xi(x)$. For the logistic
readouts, the corresponding positive-class probability is obtained by applying
the sigmoid map to the score. For the approximate-kernel probes, we use matched
feature lifts
$\varphi_{\mathrm{RFF}}, \varphi_{\mathrm{Nys}} : \mathbb{R}^{E|\mathcal{O}_n|} \to \mathbb{R}^{m}$
with projection dimension $m=1024$, so these heads test whether QKS gains
persist after a controlled nonlinear lift of fixed size before the final linear
score layer. In the final paired benchmark, we compare each QKS model against a
matched direct-readout baseline, meaning that the same shortlisted
representation vector $r(x)$ is passed directly to the same readout family
instead of first being mapped to $\xi(x)$. The direct-readout baseline and the
QKS model therefore share the same data partitions, the same input
representation, and the same readout family; only the feature map changes.

\paragraph{Protocol}
Stages~1--4 rank candidates by validation AUROC first and validation $F_1$
second. Stage~5 reports test AUROC and test $F_1$ for the finalists, with
$F_1$ computed from thresholded predictions at a fixed threshold of $0.5$.

\begin{enumerate}
    \item \textbf{Stage 1: shallow architecture search.} Fix the representation
    to the $64 \times 64$ DCT representation, set depth to $D=1$, and sweep
    $n \in \{2,4,6,8,10\}$ qubits,
    $E \in \{8,16,32,64,128,256\}$ episodes, and entanglement on/off. This
    stage identifies promising shallow operating points before any additional
    re-uploading depth is introduced.

    \item \textbf{Stage 2: depth sweep.} Take the top four Stage~1
    configurations and sweep depth
    $D \in \{1,2,4,6,8,10\}$ while keeping each candidate's Stage~1 qubit
    count, episode count, entanglement setting, and the $64 \times 64$ DCT
    representation fixed. This stage asks whether additional re-uploading depth
    improves performance once a strong shallow configuration has already been
    identified.

    \item \textbf{Stage 3: matched depth--episode trade-off.} Retain only the
    single best Stage~2 family, meaning the same representation, qubit count,
    and entanglement setting. Stage~3 does not keep the Stage~2
    winner's exact depth and episode count. Instead, it replaces that winning
    $(D,E)$ pair with the matched alternatives
    $(D,E) \in \{(1,256),(2,128),(4,64),(8,32)\}$. This comparison therefore
    tests whether the observed Stage~2 gain is attributable to depth itself or
    instead to a different allocation between circuit depth and the number of
    episodes.

    \item \textbf{Stage 4: representation comparison.} Evaluate the two
    strongest Stage~3 candidates across all ten input representations: raw input, four DCT variants, and five PCA
    variants, after which the best QKS configuration is retained for each shortlisted
    representation. This stage compares representation families under QKS
    settings already identified earlier, rather than reopening the full
    architecture search separately for every representation.

    \item \textbf{Stage 5: final full dataset benchmark.} For the final
    shortlisted representations, compare matched direct-readout baselines and
    QKS models across the same five readout families on the full raw training
    split, and report paired results on the held-out test data.
\end{enumerate}

This design keeps the combinatorial search tractable while yielding
interpretable answers to five distinct questions: which shallow architectures
are promising, whether depth helps on a fixed shallow shortlist, whether the
Stage~2 gain persists under the matched depth--episode trade-off, which input
representations remain most effective under the shortlisted QKS settings, and
whether those gains persist under a final paired benchmark against matched
direct-readout baselines.

%% file: sections/experiments.tex
\section{Results}
\label{sec:results}

\begin{figure*}[t]
    \centering
    \begin{subfigure}[t]{0.49\textwidth}
        \centering
        \includegraphics[width=\linewidth]{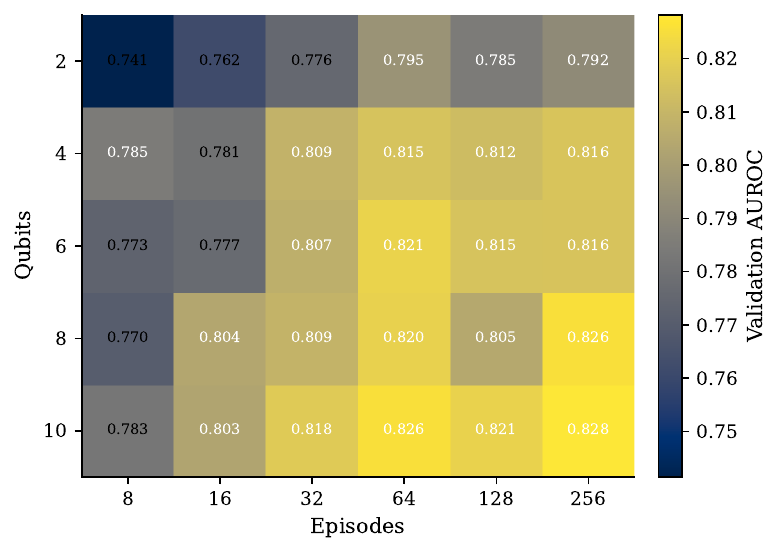}
        \caption{Entangled shallow QKS AUROC heatmap on fixed \texttt{dct64x64}.}
        \label{fig:development_stage1_entangled}
    \end{subfigure}\hfill
    \begin{subfigure}[t]{0.49\textwidth}
        \centering
        \includegraphics[width=\linewidth]{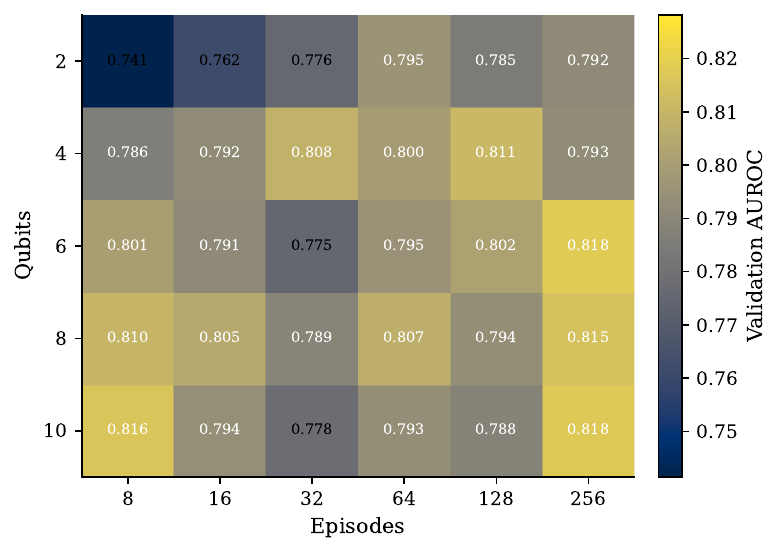}
        \caption{Unentangled shallow QKS AUROC heatmap on fixed \texttt{dct64x64}.}
        \label{fig:development_stage1_unentangled}
    \end{subfigure}

    \vspace{0.5em}

    \begin{subfigure}[t]{0.49\textwidth}
        \centering
        \includegraphics[width=\linewidth]{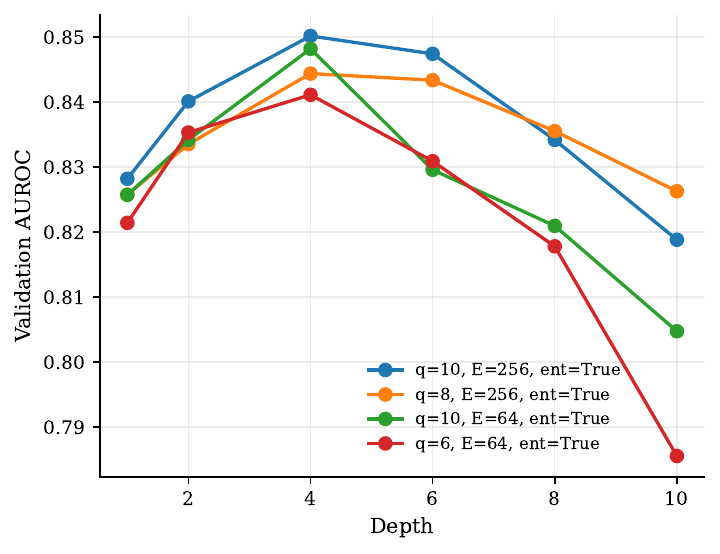}
        \caption{Depth-response curves for the four shortlisted shallow configurations.}
        \label{fig:development_stage2_depth}
    \end{subfigure}\hfill
    \begin{subfigure}[t]{0.49\textwidth}
        \centering
        \includegraphics[width=\linewidth]{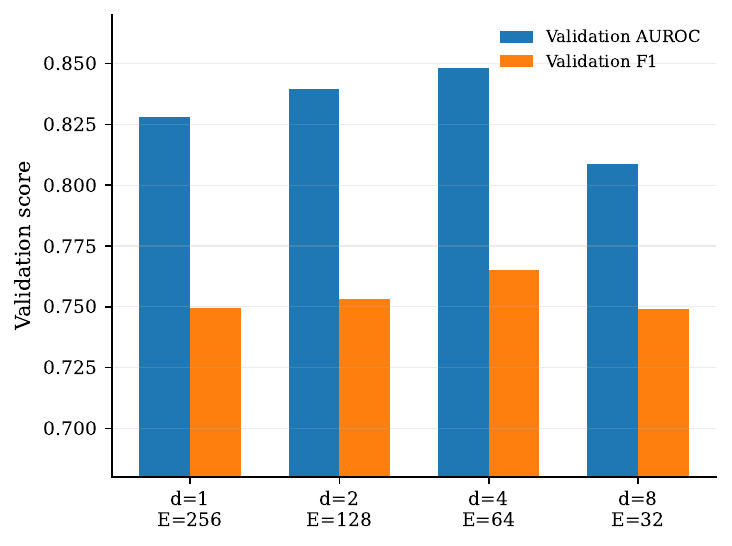}
        \caption{Validation comparison across matched $(D,E)$ pairs with fixed depth--episode allocation.}
        \label{fig:development_stage3_budget}
    \end{subfigure}
    \caption{Validation-stage coreset results used to select finalists across Stages~1--3 of the ablation protocol.}
    \label{fig:development_results}
\end{figure*}

\subsection{Ablation study of pipeline design choices}

Figure~\ref{fig:development_results} summarizes the evidence from the
model-selection stages. Its four panels isolate the entangled Stage~1 sweep ( 
Fig.~\ref{fig:development_stage1_entangled}), the unentangled Stage~1 sweep (Fig.~\ref{fig:development_stage1_unentangled}), the Stage~2 depth-response
curves (Fig.~\ref{fig:development_stage2_depth}), and the Stage~3 matched
depth--episode trade-off (Fig.~\ref{fig:development_stage3_budget}).

\paragraph{Stage 1: shallow architecture search}
On the fixed $64 \times 64$ DCT representation, the best shallow model uses
    $10$ qubits, $256$ episodes, depth $1$, and entanglement, reaching a validation
AUROC of $0.828$ and a validation $F_1$ score of $0.749$. The second largest
alternative is a $8$-qubit, 256-episode entangled configuration, which
reaches $0.826$ AUROC and $0.729$ $F_1$. Because The $10$-qubit, $256$-episode model leads on both validation AUROC and validation F1, and is therefore the Stage 1 winner under the AUROC-first, F1-second rule.
Figures~\ref{fig:development_stage1_entangled}
and~\ref{fig:development_stage1_unentangled} therefore suggest that the
performance landscape is governed less by a broad monotonic dependence on the
number of qubits or QKS episodes than by a limited set of favorable
hyperparameter combinations. In the entangled case, the best-performing
configurations remain concentrated within a relatively compact region of the
sweep, whereas the unentangled case exhibits a flatter and less structured
profile, with improvements appearing more sporadically. This suggests that,
for the fixed $64 \times 64$ DCT input, increasing the size of the randomized
feature map does not necessarily translate into better performance. Instead,
moderate configurations appear sufficient to capture much of the discriminative
structure, while larger settings may introduce redundancy and additional model-selection variance.

\paragraph{Stage 2: depth sweep}
Depth materially improves performance once a strong shallow shell has been
identified. Here, the four top performing models from Stage 1 are evaluated, and the best Stage~2 model uses $10$ qubits, $256$ episodes, depth $4$,
and entanglement, reaching $0.8502$ validation AUROC and $0.7718$ validation
$F_1$. On that same shell, moving from depth $1$ to depth $4$ raises
validation AUROC from $0.8282$ to $0.8502$ and validation $F_1$ from $0.7494$
to $0.7718$. Figure~\ref{fig:development_stage2_depth} shows a non-monotonic
depth trend across the four shortlisted shells: each curve improves once some
re-uploading is added, but the strongest entangled $q = 10$, $E=256$
configuration peaks at depth~$4$ and then declines. The benchmark therefore
favors a moderate re-uploading depth rather than the largest tested depth. This behavior is consistent with repeated data re-uploading enriching the nonlinear feature map up to a moderate depth, after which additional layers increasingly randomize or dilute task-relevant structure relative to the fixed linear readout.
The fact that the strongest configurations are consistently entangled further suggests that interaction terms between qubits provide useful additional structure beyond independent single-qubit transformations. Overall, the results indicate that the best performance is obtained by combining entanglement with a moderate re-uploading depth, while increasing circuit depth beyond this regime provides no systematic benefit and  may degrade generalization.

\paragraph{Stage 3: matched depth--episode trade-off}
Stage~3 introduces a fixed-budget comparison in which the nominal product
$D \times E = 256$ is held constant while that budget is redistributed between
circuit depth and the number of episodes. It remains directly anchored to
Stage~2 in that it keeps the winning Stage~2 family fixed in terms of
representation, qubit count, and entanglement, but it does \emph{not} preserve
the Stage~2 winner's exact $(D,E)=(4,256)$ setting. Instead, it evaluates the
matched alternatives $(D,E)\in\{(1,256),(2,128),(4,64),(8,32)\}$ in order to
test whether the Stage~2 gain is genuinely attributable to depth or whether a
similar budget is better spent on additional episodes. The strongest candidate is the $10$-qubit, $64$-episode, depth-$4$ model, with $0.8482$ validation AUROC and $0.7649$ validation $F_1$. The $10$-qubit, $128$-episode, depth-$2$
alternative remains competitive at $0.8397$ AUROC and $0.7531$, while
the shallow $(D,E)=(1,256)$ and deep $(D,E)=(8,32)$ endpoints are weaker.
Figure~\ref{fig:development_stage3_budget} makes the trade-off explicit: within
this fixed-allocation comparison, $(D,E)=(4,64)$ leads on both validation
AUROC and validation $F_1$, $(D,E)=(2,128)$ remains the strongest alternative,
and the shallow and very deep endpoints trail. The Stage~3 result suggests that the best performance comes from balancing per-episode expressivity and cross-episode diversity: too little depth leaves each episode under-expressive, whereas too few episodes reduce the averaging benefits of the randomized ensemble.

\begin{figure*}[t]
    \centering
    \includegraphics[width=0.84\textwidth]{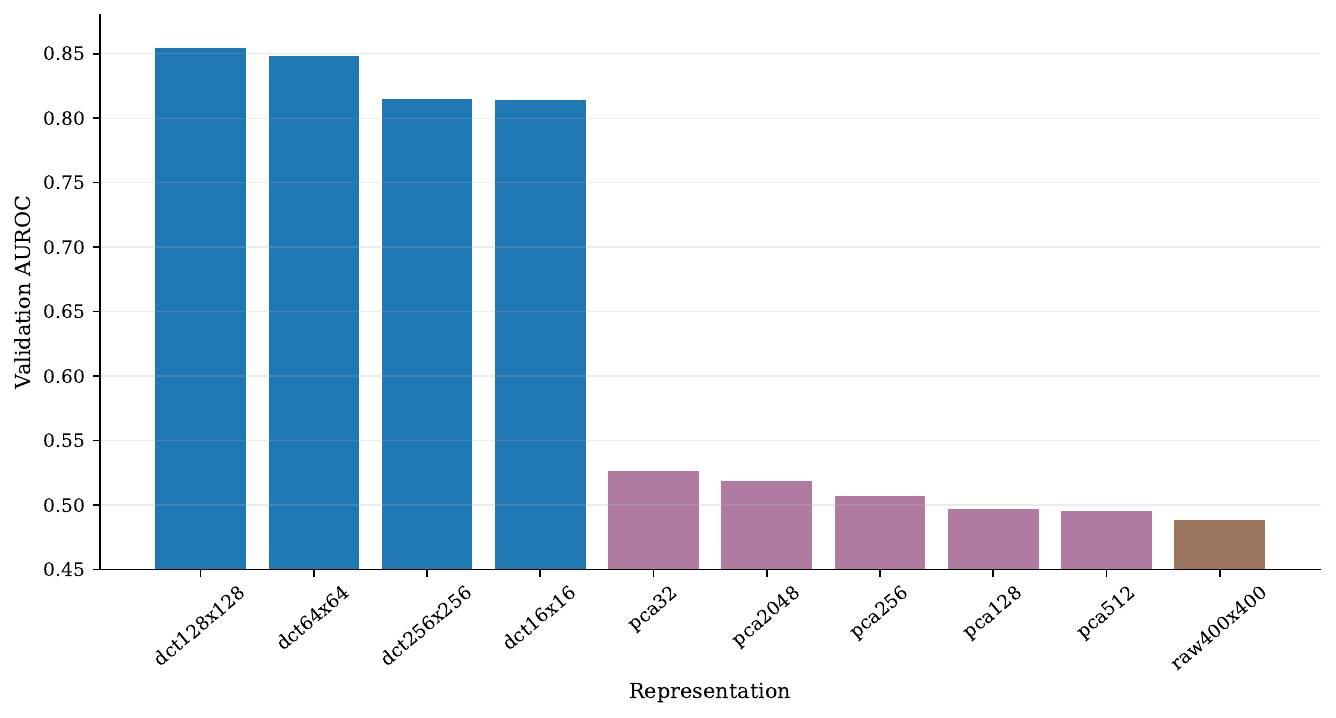}
    \caption{Stage~4 comparison across input representations. Each bar reports
    the validation AUROC obtained by applying the two strongest Stage~3 QKS
    candidates across the ten representation families. DCT representations
    dominate, while PCA and raw inputs remain well below the advancing cutoff.}
    \label{fig:stage4_transfer}
\end{figure*}

\paragraph{Stage 4: representation comparison}
When the two strongest Stage~3 QKS candidates are compared across all suggested input representations, the DCT family separates sharply from the
remaining ones. The three strongest advancing representations are
 \texttt{dct128x128}, \texttt{dct64x64},and \texttt{dct256x256}; their best QKS
configurations all remain in the entangled $10$-qubit regime. The best non-DCT
contender, \texttt{pca32}, reaches only $0.5262$ validation AUROC, well below
the DCT advancing cutoff. Figure~\ref{fig:stage4_transfer} makes this pattern
explicit across all ten input representations, while Table~\ref{tab:stage4_shortlist}
reports the three DCT finalists that advance to Stage~5, where the final QKS
settings are fixed and for the final evaluation. Under this comparison, the benchmark does not support a
representation-agnostic QKS benefit; instead, it identifies a DCT-dominant
regime in which structure-preserving classical preprocessing remains part of
the effective hybrid model. This sharp DCT advantage is consistent with the idea that anomaly signatures are organized in the time--frequency plane and are therefore better preserved by low-index DCT coefficients than by raw flattened inputs or PCA directions learned from global variance.

\begin{table}[t]
    \centering
    \small
    \caption{Stage~4 finalists that advance under the representation
    comparison and their locked QKS configurations.}
    \label{tab:stage4_shortlist}
    \begin{tabular}{lrrrrcc}
        \toprule
        Repr. & $q$ & Ep. & $D$ & Ent. & Val AUROC & Val $F_1$ \\
        \midrule
        \texttt{dct128x128}  & 10 &  64 & 4 & Y & 0.8541 & 0.7774 \\
        \texttt{dct64x64} & 10 & 64 & 4 & Y & 0.8482 & 0.7649 \\
        \texttt{dct256x256}  & 10 &  128 & 2 & Y & 0.8152 & 0.7299 \\
        \bottomrule
    \end{tabular}
\end{table}
\begin{table*}[t]
    \centering
    \small
    \caption{Compact Stage~5 benchmark slice for the linear readouts.}
    \label{tab:final_linear_benchmark}
    \begin{tabular}{llrrrrrr}
        \toprule
        Repr. & Readout & Direct AUROC & QKS AUROC & $\Delta$AUROC & Direct $F_1$ & QKS $F_1$ & $\Delta F_1$ \\
        \midrule
       \texttt{dct64x64} & \texttt{LinearSvm}      & 0.6939 & 0.8778 & 0.1839 & 0.6515 & 0.7938 & 0.1423 \\
        \texttt{dct64x64} & \texttt{LinearLogistic} & 0.7266 & 0.8717 & 0.1451 & 0.6607 & 0.7995 & 0.1389  \\
        \texttt{dct128x128} & \texttt{LinearSvm}      & 0.6945 & 0.8763 & 0.1818 & 0.6519 & 0.7969 & 0.1451 \\
        \texttt{dct128x128} & \texttt{LinearLogistic} & 0.7294 & 0.8705 & 0.1411 & 0.6614 & 0.7978 & 0.1363 \\
        \texttt{dct256x256} & \texttt{LinearSvm}      & 0.6975 & 0.8576 & 0.1601& 0.6558 & 0.7754 & 0.1195 \\
        \texttt{dct256x256} & \texttt{LinearLogistic} & 0.7312 & 0.8264 & 0.0952 & 0.6635 & 0.7739 & 0.1104 \\
        \bottomrule
    \end{tabular}
    
\end{table*}
\subsection{Final benchmark on the full dataset}
\begin{figure*}[t]
    \centering
    \includegraphics[width=0.82\textwidth]{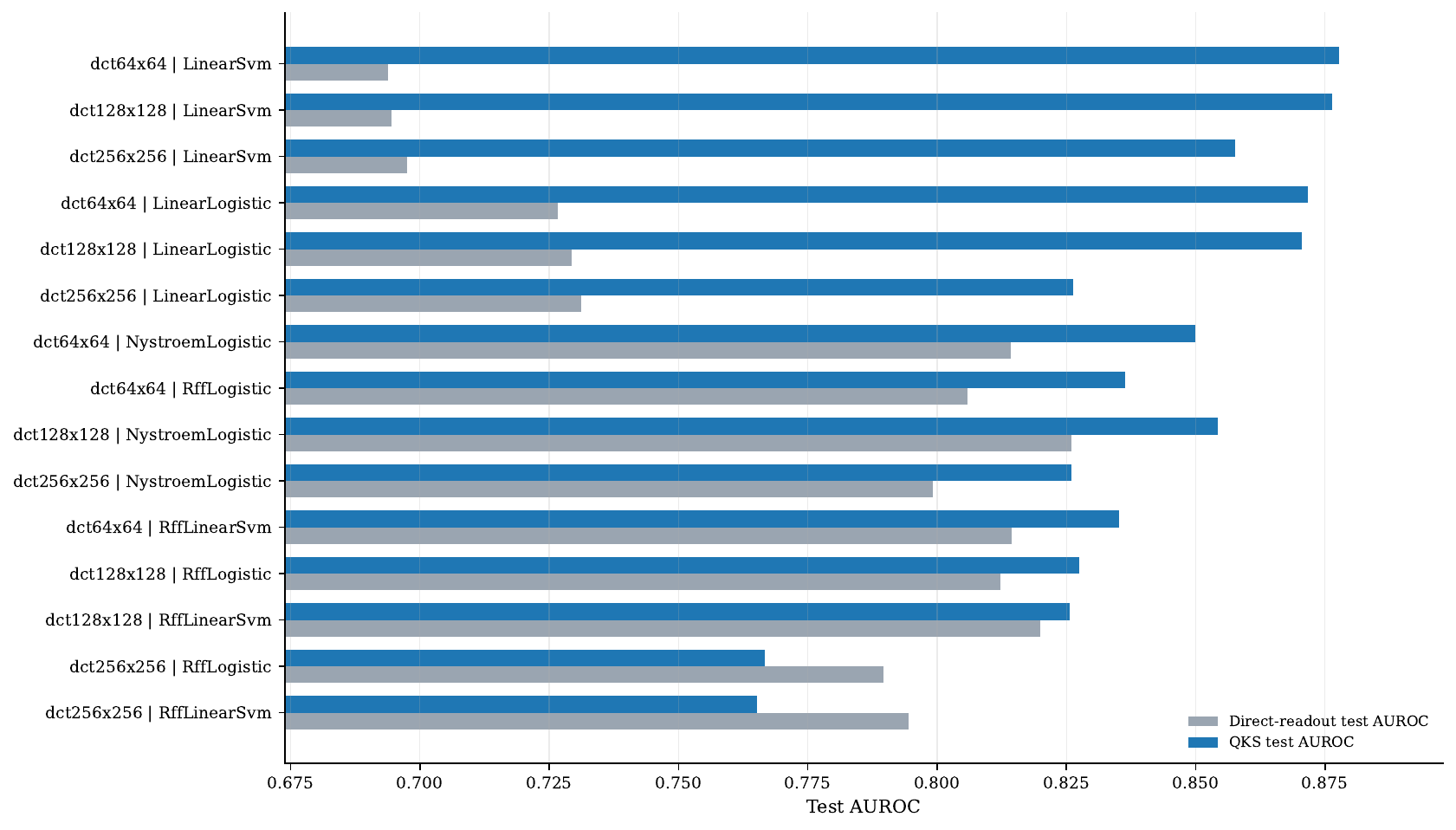}
    \caption{Stage~5 paired test AUROC comparison between matched
    direct-readout baselines and QKS pipelines. Each row corresponds to one
    representation--readout comparison on the untouched raw test split after
    the locked finalists from a single finalized benchmark run are retrained on
    the full raw training split. The largest AUROC gaps appear for the linear
    readouts, while the approximate-kernel readouts also benefit from QKS but
    by smaller margins.}
    \label{fig:final_benchmark}
\end{figure*}

Stage~5 retrains the Stage~4 finalists on the full raw training split and
evaluates them once on the held-out raw test split. Among the $15$ final
representation--readout comparisons, the highest test AUROC is obtained by
\texttt{dct64x64} + LinearSvm, which reaches $0.8778$ test AUROC and $0.7938$
test $F_1$. The highest test $F_1$ is obtained by
\texttt{dct64x64} + LinearLogistic, which reaches $0.7995$ at a test AUROC of
$0.8717$. Figure~\ref{fig:final_benchmark} summarizes the paired
direct-readout-versus-QKS test AUROC comparison across all $15$
representation--readout combinations, while
Table~\ref{tab:final_linear_benchmark} reports the exact Stage~5 numbers for
the six linear cases that exhibit the largest gains. The fact that the best
AUROC and best $F_1$ arise from different DCT configurations suggests that
retaining more coefficients is most useful for ranking quality, whereas a more
compact representation can yield a cleaner decision boundary at the fixed
threshold used for $F_1$.

Two factual patterns can be identified, with their interpretation deferred to
Section~\ref{sec:discussion}. First, all $15$ rows in
Fig.~\ref{fig:final_benchmark} place the QKS pipeline above its matched
direct-readout baseline, so the AUROC gain is systematic across the finalized
benchmark rather than confined to a single representation--readout pair. Second,
writing $\Delta$AUROC and $\Delta F_1$ for the QKS score minus the corresponding
direct-readout baseline score, the largest gains arise for the linear readouts:
across the six linear rows in Table~\ref{tab:final_linear_benchmark} the AUROC
gains range from $+0.0952$ to $+0.1839$, with the largest gap attained by
\texttt{dct64x64 + LinearSvm}, while the approximate-kernel readouts improve by
visibly smaller margins.

\subsection{Validation on real quantum hardware}
\begin{table*}[t]
\centering
\caption{Real-QPU versus simulator performance across classical readouts, under identical settings.}
\begin{tabular}{lrrrrrr}
\toprule
\textbf{Readout} & \textbf{Hardware AUROC} & \textbf{Hardware F1} & \textbf{Simulator AUROC} & \textbf{Simulator F1} & \textbf{$\Delta_{HS}$ AUROC} & \textbf{$\Delta_{HS}$ F1} \\
\midrule
\texttt{LinearSvm}         & 0.738302 & 0.605144 & 0.731170 & 0.657754 & $+7.132\times10^{-3}$ & $-5.261\times10^{-2}$ \\
\texttt{LinearLogistic}    & 0.720810 & 0.658851 & 0.733619 & 0.645161 & $-1.281\times10^{-2}$ & $+1.369\times10^{-2}$ \\
\texttt{RffLinearSvm}    & 0.807438 & 0.719892 & 0.805798 & 0.704762 & $+1.640\times10^{-3}$ & $+1.513\times10^{-2}$ \\
\texttt{RffLogistic}       & 0.803232 & 0.736967 & 0.805961 & 0.702247 & $-2.729\times10^{-3}$ & $+3.472\times10^{-2}$ \\
\texttt{NystroemLogistic}  & 0.813637 & 0.748166 & 0.815457 & 0.716418 & $-1.819\times10^{-3}$ & $+3.175\times10^{-2}$ \\
\bottomrule
\end{tabular}

\label{tab:ibm_sim_readout_comparison}
\end{table*}

% This experiment completes the two-level realism of the study: 
% the dataset is grounded in real measured sub-6,GHz cellular signal, and the pipeline is here validated on real quantum hardware, with anomaly generation remaining deliberately synthetic and controlled\ola{Is this paragraph necessary here (I mean talking about the overall setting including data)? if we lack space, it can be omitted}.
To assess the hardware-realism aspect of the study, we performed a focused validation on quantum hardware. Due to the cost of real-device execution, all experiments were conducted with an entangled feature map using 4 qubits, a single-layer depth, and 32 episodes, with expectation values estimated from 4096 shots per circuit. These experiments were run on the \textit{ibm\_quebec} QPU from 25\textsuperscript{th} to 29\textsuperscript{th} March 2026, using the \texttt{dct128x128} representation. In contrast to the completed simulation study, the measured observable family was restricted to all single-qubit \(Z\) observables and pairwise \(ZZ\) observables,
\begin{equation}
\mathcal{O}_{Z,ZZ}^{(n)}
=
\{\sigma_Z^{(i)}\}_{i=0}^{n-1}
\cup
\{\sigma_Z^{(i)}\sigma_Z^{(j)}\}_{0 \le i < j < n},
\end{equation}
so that $|\mathcal{O}_{Z,ZZ}^{(n)}| = \frac{n(n+1)}{2},$
instead of the full one- and two-body Pauli family considered in simulation. In addition, the dataset size was reduced to approximately 10\% of the original training and test sets. The search space was deliberately restricted so as to remain within the available quantum budget and to limit the practical overhead associated with queue times and repeated hardware calls. As a result, this experiment was not intended to reproduce the full simulation study, but rather to evaluate the practical viability of the approach under genuine real-device conditions. It should therefore be viewed as a focused real-QPU benchmark aimed at assessing whether the learned representation retains meaningful predictive structure when executed on an actual superconducting quantum processor.

The resulting comparison between real-QPU and simulator performance is reported in Table~\ref{tab:ibm_sim_readout_comparison} for several classical readout models, with the simulator baseline obtained by re-running the pipeline under exactly the same configuration used on real-QPU---identical feature map, number of qubits, circuit depth, number of episodes, observable family $\mathcal{O}_{Z,ZZ}^{(n)}$, and train/test splits---so that the execution backend is the only varying factor.
Overall, the real-hardware results remain close to their simulated counterparts, with only small deviations in AUROC across all tested readouts. Here, HS denotes the hardware--simulator difference:
\[
\Delta_{\mathrm{HS}} \mathrm{AUROC}= \mathrm{AUROC}_{\mathrm{QPU}} - \mathrm{AUROC}_{\mathrm{sim}}.
\]
More specifically, the observed AUROC gaps range from \(-1.281 \times 10^{-2}\) to \(+7.132 \times 10^{-3}\), indicating that the ranking quality of the QKS features is largely preserved under real-device execution despite the reduced measurement setting. The F1-score differences,
\[
\Delta_{\mathrm{HS}} \mathrm{F1} = \mathrm{F1}_{\mathrm{QPU}} - \mathrm{F1}_{\mathrm{sim}},
\]
are somewhat more variable, but remain favorable for most readouts. In particular, the \texttt{NystroemLogistic} readout achieves the best real-QPU performance, with an AUROC of 0.813637 and an F1 score of 0.748166, compared with 0.815457 and 0.716418 in simulation. Taken together, these results indicate that, within this deliberately reduced experimental setting, the proposed pipeline remains robust to real-device execution and preserves a substantial fraction of the predictive structure observed in simulation.

\section{Discussion and Limitations}
\label{sec:discussion}

\paragraph{Multi-depth QKS acts as a representation enhancer}
The completed benchmark does not indicate that QKS uniformly outperforms classical baselines. Rather, repeated data re-uploading improves separability in specific representation regimes, with the largest gains obtained at moderate depths, beyond the shallow template but below the highest tested values. Entanglement also contributes in the strongest regimes: entangled configurations dominate the top shallow models and outperform matched unentangled variants in most comparisons, although not systematically across the full search space. Overall, the best QKS representations result from the combination of a suitable input representation, moderate re-uploading depth, and useful inter-qubit interactions, rather than from simply increasing circuit complexity.

\paragraph{The strongest regime is DCT-specific}
The feature-representation stage is decisive: all top-performing configurations use DCT variants, whereas PCA and raw inputs do not remain competitive. This suggests that, for spectrogram-based RF anomaly detection, performance depends not on the QKS feature map alone, but on its combination with a suitable classical representation. Our interpretation is that DCT is more effective because it preserves the structured organization of anomaly patterns in the time--frequency plane while concentrating informative content into a compact set of coefficients. By contrast, PCA emphasizes directions of maximal global variance, which can mix distant time--frequency regions and obscure features that are more relevant for anomaly discrimination. Raw inputs, in turn, likely preserve excessive irrelevant variability and redundancy, thereby reducing the ability of the QKS feature map to produce a representation aligned with the discriminative structure of the task.

\paragraph{Readout dependence remains informative}
All final readouts benefit from QKS, yet the largest paired gains appear for the
linear probes. This is informative as it
shows that QKS is changing the geometry seen by simple classical classifiers,
not merely adding redundancy that only richer downstream models can exploit. In
practical terms, this is the main value of the QKS features in this study: they
act as a representation enhancer whose effect is largest when the downstream
classifier is deliberately simple, while still yielding a positive, but smaller,
lift for the nonlinear readouts.

\paragraph{Real-device execution preserves feature quality}
The hardware study confirms that QKS features retain meaningful 
predictive structure under real-device conditions. Although the 
QPU experiment uses a reduced configuration with fewer qubits, 
shallower depth, a restricted observable set, and a smaller dataset, 
the results remain within 0.013 AUROC of their simulated counterparts 
across all tested readouts. Notably, the ranking order across readouts 
is preserved between hardware and simulation, with 
\texttt{NystroemLogistic} achieving the best performance in both 
settings. This consistency indicates that the feature geometry learned 
by the QKS pipeline is not critically disrupted by hardware noise at 
this scale, and supports the practical viability of the approach on 
near-term superconducting processors. We emphasize that this experiment 
was not designed to reproduce the full simulation benchmark, but to 
assess whether the pipeline remains functional under genuine real-device 
constraints, a question it answers affirmatively. Taken together, these observations suggest that the study probes realism along both the data and hardware axes, placing the benchmark closer to realistic hybrid deployment conditions.

%\AAcomment{[Place here with a different title discussion about real-QPU results.]}

\paragraph{Limitations}
The evidence remains task-specific. The dataset is controlled rather than drawn
from a live operational RF deployment, and generalization to unseen interference types or real-world spectrum conditions has not been tested.
The completed study uses one finalized run rather than a broad seed ensemble, so 
stability across random initializations of the affine maps has 
not been quantified. The classical readout family is intentionally restricted to fast probes and kernel approximations, and stronger classical heads may reduce or eliminate the observed QKS gains. We  therefore make no universal claim about QKS on every anomaly detection task, and we do not interpret the present  results as evidence of quantum advantage.

%% file: sections/conclusion.tex
\section{Conclusion}
\label{sec:conclusion}

This paper studied Quantum Kitchen Sinks for RF spectrogram anomaly detection under a controlled, leakage-free evaluation protocol. Starting from the standard shallow QKS template, we introduced a multi-depth variant with repeated data re-uploading and optional ring entanglement, and evaluated it on spectrograms built from real measured sub-6\,GHz cellular signals with synthetically injected anomalous transmissions. The resulting benchmark was designed not only to compare models, but also to identify which components of the hybrid pipeline matter most for this task.

The results support a clear picture within this benchmark. QKS does not provide a representation-agnostic improvement over classical baselines; instead, its utility depends strongly on the interaction between the quantum feature map, the input representation, and the downstream readout. The strongest results arise in DCT-based regimes, where moderate-depth entangled QKS configurations outperform their matched direct-readout baselines across all evaluated final readout families on held-out test data, reaching a best test AUROC of 0.8778 and a best test $F_1$ of 0.7995. The same favorable pattern is not reproduced with raw or PCA-based inputs, and the largest gains do not come from the largest tested models. In addition, the reduced real-device validation on the \textit{ibm\_quebec} QPU preserves the main predictive trends seen in simulation, with AUROC deviations below 0.013. Future work should extend the study to additional RF anomaly families and test whether the same operating regimes persist under broader hardware-aware execution constraints.

%\ola{This Section must be added after paper acceptance}
\section{Acknowledgement}
\label{sec:ack}
The authors gratefully acknowledge the financial support provided by Defence Research and Development Canada (DRDC), which made this research possible.